\documentclass{esannV2}
\usepackage[dvips]{graphicx}
\usepackage[latin1]{inputenc}
\usepackage{amssymb,amsmath,array}
\usepackage[skip=2pt]{caption}
\usepackage{subcaption}
\usepackage{bmpsize}
\usepackage{float}
\usepackage{url}
\usepackage{array, booktabs, makecell}

\begin{document}
\title{Machine learning for automated quality control in injection moulding manufacturing}

\author{Steven Michiels$^1$, C\'edric De Schryver$^2$, Lynn Houthuys$^1$,\\ Frederik Vogeler$^1$, Frederik Desplentere$^2$
%
\thanks{The research leading to these results has received funding from VLAIO (TETRA project 'AI4IM', project number HBC.2020.2556)}
%
\vspace{.3cm}\\
%
1- Thomas More University of Applied Sciences, Belgium  \\
Dept. of Smart Technology \& Design, Campus De Nayer
%
\vspace{.1cm}\\
2- KU Leuven - University of Leuven, Belgium \\
Dept. of Materials Engineering, ProPoliS Research Group, Campus Bruges
}

\maketitle

\begin{abstract}
Machine learning (ML) may improve and automate quality control (QC) in injection moulding manufacturing. As the labelling of extensive, real-world process data is costly, however, the use of simulated process data may offer a first step towards a successful implementation. In this study, simulated data was used to develop a predictive model for the product quality of an injection moulded sorting container. The achieved accuracy, specificity and sensitivity on the test set was $99.4\%$, $99.7\%$ and $94.7\%$, respectively. This study thus shows the potential of ML towards automated QC in injection moulding and encourages the extension to ML models trained on real-world data.
\end{abstract}

\section{Introduction}

When it comes to producing large series of plastic products, injection moulding may be the most important production technique for the manufacturing industry. The large batch sizes, however, result in a time-intensive and therefore costly quality control (QC). At the same time, large amounts of data are generated during the manufacturing process, e.g. pressure and temperature signals from the machine itself or from within the mould. Harnessing these data through machine learning (ML) may improve and automate the QC, hereby potentially reducing cost while ensuring high product quality \cite{Weichert2019, Fernandes2018}.

ML techniques have been successfully applied to improve the injection moulding process in various ways. For instance, neural networks (NN) have been used to predict the melt temperature during plastification \cite{Librantz2017} or to predict the roughness of a plastic mould \cite{Zhao1999}. Neural networks have also been used for automated QC, for instance to predict various quality criteria \cite{Lopes2000b} or to predict product weight as a quality indicator \cite{Chen2008}. Both approaches use various process parameters as input. Another approach created extra input parameters by applying a thermal camera to the end-products and subsequently combining a convolutional NN and long short-term memory to assess product quality \cite{Nagorny2017}.

In a real-world setting, however, labelling large datasets can be very expensive. A more cost-effective approach may be to build a model on simulated data and then finetune the model on real-world data via transfer learning \cite{Tercan2018}.  In this paper, such an approach was investigated, by using simulated data for important physical parameters during the injection moulding process. Hence, in contrast to the work mentioned above, no process parameters or additional measurement equipment was used. A ML model to predict the product quality of a small sorting container, based on its dimensions, was successfully developed. The model was fully trained and tested on simulated data, in order to be validated on and transferred to real-world data in a later stage.


\begin{figure}
     \centering
     \begin{subfigure}[b]{0.3\textwidth}
         \centering
         \includegraphics[width=\textwidth]{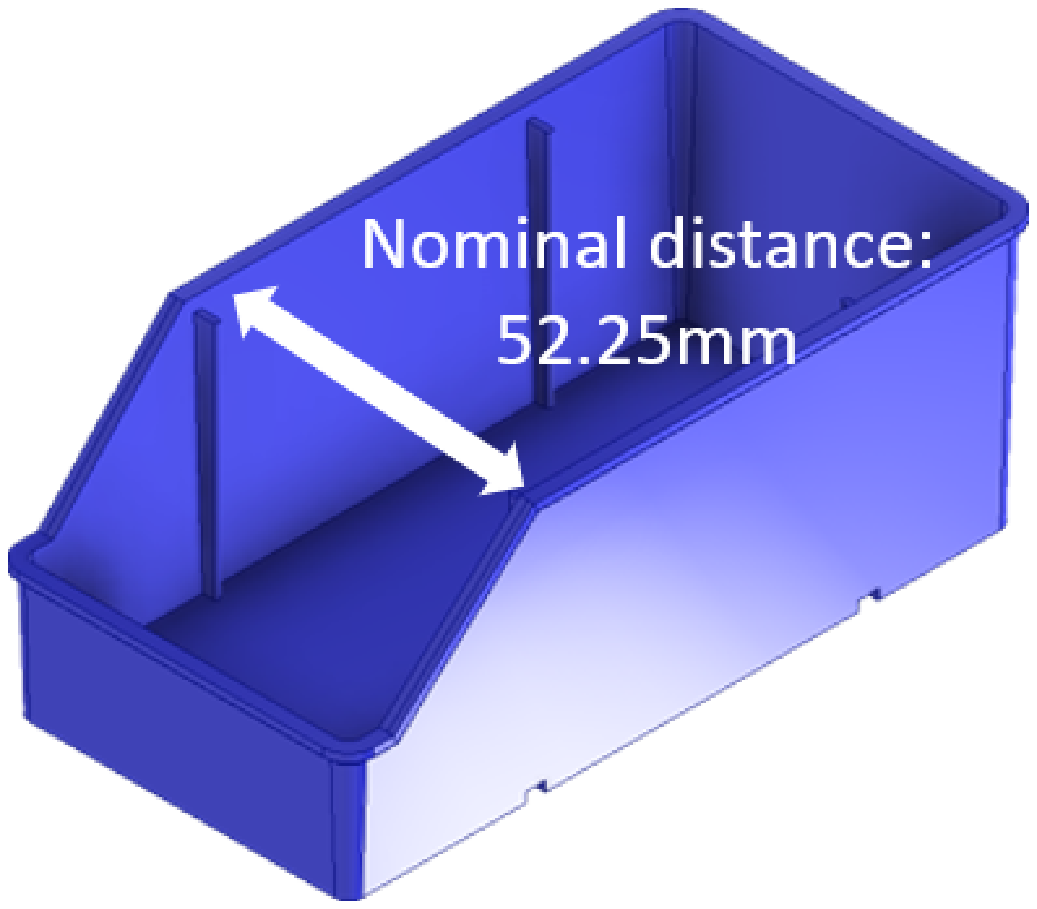}
         \caption{}
         \label{fig:Sorteerbakje}
     \end{subfigure}
     \quad \quad \quad \quad
		 \begin{subfigure}[b]{0.4\textwidth}
         \centering
         \includegraphics[width=\textwidth]{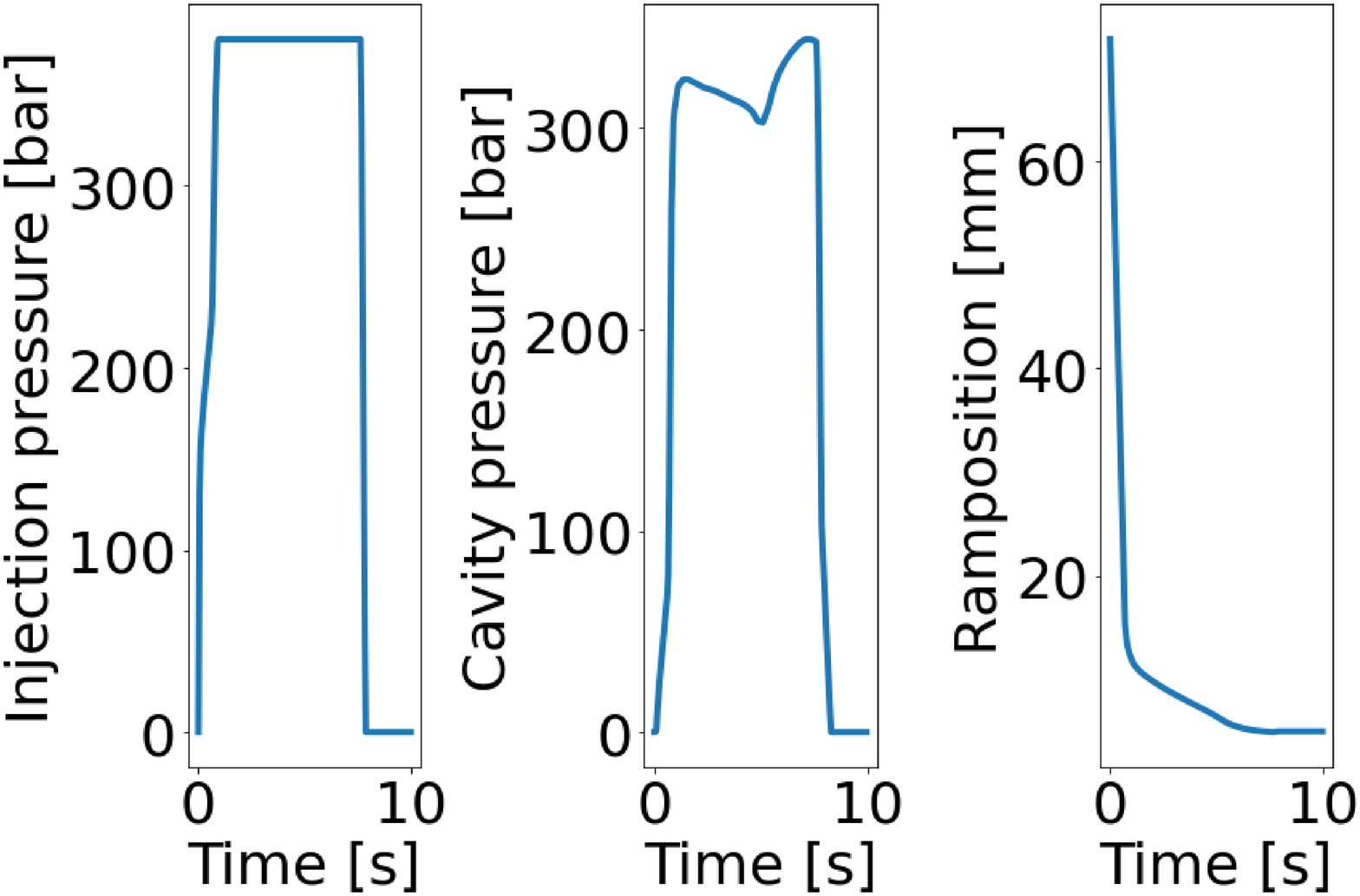}
         \caption{}
         \label{fig:Timeseries}
     \end{subfigure}
     
        \caption{a) The injection moulded object considered in this study, i.e. a sorting container. The nominal distance between the upright planes, i.e. the 'opening distance', was the considered metric for the quality control. b) Time series for the physical parameters, resulting from one example Moldflow simulation.}
        \label{fig:three graphs}
\end{figure}




\section{Materials \& Methods}
\subsection{Data generation using simulations}
Figure \ref{fig:Sorteerbakje} shows the object considered in this study, i.e. a plastic sorting container. The injection moulding of the object was thermo-mechanically modeled using moulding process-specific software (Moldflow 2021, Autodesk). The polymer used in the simulations was Sabic PHC27, a standard grade unfilled polypropylene (PP). The software was coupled with MATLAB (version R2019b, MathWorks), in order to perform simulations in batch while sequentially varying relevant process parameters within practically relevant ranges. These parameters namely influence the shrinkage of the polymer object and therefore determine the resulting dimensions of the manufactured object.\\The used input parameters and their variation ranges are displayed in Table \ref{table:inputparams}. The output of the simulations is summarized in Table \ref{table:outputs}. Each simulation  yielded time series for important physical parameters during the injection moulding process, i.e. the cavity pressure, cavity temperature and ramposition, as shown in Figure \ref{fig:Timeseries}. Each simulation also yielded the resulting dimensions of the virtually manufactured object. The considered metric for the QC of the object was the distance between the upright planes of the object, i.e. the 'opening distance', as shown in Figure \ref{fig:Sorteerbakje}. The population mean $\mu$ and standard deviation $\sigma$ of this metric were estimated by calculating the mean and standard deviation of the opening distance over the whole dataset. A manufactured product was then assigned to a positive (or 'rejected') quality class if the opening distance deviated more than $2\sigma$ from $\mu$. The goal of the machine learning model was to predict the quality class for unseen objects.  

\begin{table}
\centering 
\begin{tabular}{c c c} 
Varied input parameter & Median & Interquartile range\\ 
\hline 
Cooling water temperature $[K]$ & $313.4$ & $310.1-316.5$ \\ 
Melt temperature $[K]$ & $503.2$  & $500.8-505.4$ \\
Flowrate $[cm^3/s]$ & $40.0$ & $38.1-41.9$\\
Packing pressure $[\text{bar}]$ & $400.1$  & $381.8-417.9$ \\
Power law index $n$ in viscosity model $[-]$ & $0.23$  & $0.21-0.25$ \\
D1-constant in viscosity model $[Pa \cdot s]$ & $5.8 E13$ & $5.6-6.0 E13$ \\

\hline 
\end{tabular}
\caption{Input variations for the injection moulding simulations. In order to incorporate variations in the rheological properties as well, the n-value (Power law index) and the data-fitted D1-value of the Cross-WLF viscosity model were adjusted within the given ranges \cite{autodesk}. }
\label{table:inputparams} 
\end{table}

\begin{table}
\centering 
\begin{tabular}{c c} 
Output & Purpose\\ 
\hline 
Time series for injection pressure & Input for automated feature extraction  \\ 
Time series for cavity pressure & Input for automated feature extraction  \\
Time series for ramposition & Input for automated feature extraction  \\
\hline 
Opening distance & Target variable for ML model  \\
Quality class & Target variable for ML model\\

\hline 
\end{tabular}
\caption{Output of the injection moulding simulations.}
\label{table:outputs} 
\end{table}

\subsection{Data processing, feature engineering and data splitting}
All data processing, feature engineering and model training was performed in Python (version 3.8). Feature extraction from the time series as generated by the simulations was performed using the 'tshfresh' open source Python library \cite{Christ2018}. In brief, this library allows to automatically, through features, characterize the time series with respect to the distribution of the data points, correlation properties, stationarity, entropy and non-linear time series analysis. As such, 3156 features were extracted per observation (= per moulded object), for a total of 3147 observations. The feature set was reduced by only retaining the 300 features with the largest correlation (in absolute value) with the opening distance and the quality class, respectively. The quality class exhibited severe class imbalance, i.e. a 20:1 ratio between the positive (i.e. 'accepted') and negative (i.e. 'rejected') class, respectively. Using stratified shuffle splitting based on the quality class, the observations and their correlation-filtered features were split in a training set ($80\%$ of the data), a test set ($10\%$) and a hold-out set ($10\%$) for future use.

\subsection{Model training and testing}
\begin{figure}[H]%
\centering
\includegraphics[width=0.75\textwidth]{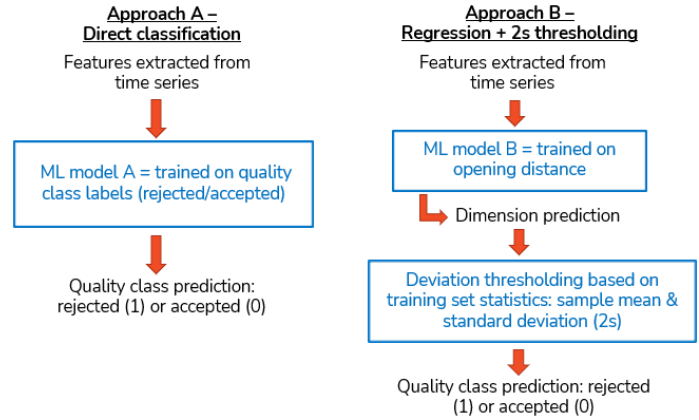}%
\captionsetup{justification=centering}
\caption{Two different approaches were elaborated: a direct classification approach and a regression approach with subsequent thresholding based on the sample (= training set) mean $\bar{x}$ and sample standard deviation $s$.}
\label{fig:Approach}%
\end{figure}

Figure \ref{fig:Approach} shows the two different approaches that were used to create a predictive model for the quality class label. In approach A, a classification model was trained to directly predict the quality class label. Weighted classes were used to account for the class imbalance. In approach B, a regression model was trained to predict the opening distance. The quality class label was then determined based on the position of the predicted opening distance with respect to the acceptation boundaries $[\bar{x}-2s;\bar{x}+2s]$, where $\bar{x}$ is the mean opening distance over the training set and where $s$ is the standard deviation of the opening distance over the training set. To benchmark both approaches, a naive model was created that always predicted the majority class (label 0, i.e. 'accepted). \\For the non-naive model approaches, a pipeline was created that contained a standard scaler and a Light Gradient Boosting Machine (LGBM \cite{Guolin2017}) classifier or regressor, respectively. Model training was performed using 5-fold cross-validation on the training set. Hyperparameter tuning was performed using randomized search within predefined search lists for the following parameters: n\_estimators, alpha, lambda, subsample, child\_weight, child\_samples, num\_leaves \cite{LGBMClassReg}. Log loss and mean absolute error were used as cross-validation score for the classification approach (A) and regression approach (B), respectively.\\The parameter combinations that yielded the best cross-validation scores were used to retrain the model on the whole training set. Lastly, the finally obtained models were used to predict the quality class labels for the test set.

\begin{table}[ht]
\renewcommand{\arraystretch}{0.7} 
\setlength{\tabcolsep}{3pt} 
\centering 
\begin{tabular}{c c c c} 
\hspace{-10pt}Approach & \hspace{-10pt}  Accuracy $[\%]$ & Specificity $[\%]$ & Sensitivity $[\%]$  \\ 
\hline 
\hspace{-10pt}Direct classification (LGBM) & \hspace{-10pt} $99.4$ & $100.0$ & $90.5$ \\
\hspace{-10pt}Regression (LGBM) $+ 2s$ thresholding & \hspace{-10pt}  $99.4$ & $99.7$ & $94.7$ \\
\hspace{-10pt}Naive model benchmark & \hspace{-10pt}  $93.9$ & $100.0$ & $0.0$ \\
\hline 
\end{tabular}
\caption{Classification performance metrics for the different approaches evaluated on the test set.}
\label{table:performance} 
\end{table}

\begin{table}[ht]
\centering 
\begin{tabular}{c c c} 
& True positive count & True negative count\\
\hline 
Predicted positive count  & 18 & 1 \\
Predicted negative count & 1 & 292 \\
\hline 
\end{tabular}
\caption{Confusion matrix for the approach that performed best on the test set, i.e. regression with subsequent 2s thresholding (approach B). A positive class corresponded with a to be rejected item. The test set consisted of 312 observations, of which 19 were true positives.}
\label{table:confusion} 
\end{table}

\section{Results \& Discussion}
Table \ref{table:performance} shows the accuracy, specificity (= true negative rate or TNR) and sensitivity (= true positive rate or TPR) for the ML approaches A and B, as well as for the naive benchmark approach. While, by its nature, the naive model performed perfect in terms of accepting true negative objects, it failed to detect any true positive object (i.e. an object with unacceptable dimensions), although this is of crucial importance for QC of manufacturing. In this regard, the ML approaches A and B showed a marked improvement, with a sensitivity of at least $90\%$. The regression model + $2s$ thresholding (approach B) could be considered the best performing model, due to the highest sensitivity at a negligible (0.3 $\%$ points) cost of decreased specificity. Table \ref{table:confusion} shows the confusion matrix for the predictions on the test set using approach B. For a test set of 312 objects, the model classified 292 out of the 293 to be accepted (= negative) objects as such, whilst the model classified 18 out of the 19 to be rejected (= positive) objects as such.\\For potential practical applicaton of the developed model, it should be weighed whether the model can complement or replace classical QC via sampling. In this regard, the achieved TNR and TPR should be compared with the TNR and TPR achieved by QC via sampling. An additional consideration may be the cost balance between the TNR on one hand, which impacts the time and hence cost spent on additional control procedures, and the TPR, which impacts the number of products of inferior quality received by the customers. The ML model may be tuned towards achieving the pursued balance between both. Moreover, the performance of the ML model may be further increased by training on a larger dataset, especially a dataset with an improved class balance.\\Practical application of the developed model also requires careful matching of the simulation environment with the real-world injection moulding process. In this regard, the used simulation software is validated and well established to model the process. In addition, the physical process parameters of which the features were extracted, effectively can be monitored on real injection moulding installations using commercially available hardware. Finally, these real-world data could be used to enhance the developed model via transfer learning \cite{Tercan2018} and hence improve its applicability in real-world conditions. In any case, the presented study has shown the potential of ML towards automated quality control in injection moulding manufacturing.

\section{Conclusion}
A machine learning model with excellent performance in assessing product quality of simulated injection moulded objects was developed. This study therefore shows the potential of machine learning towards automated quality control in injection moulding and encourages the extension to models trained on real-world data.


\begin{footnotesize}



\bibliographystyle{unsrt}
\bibliography{AI4IMbib}

\end{footnotesize}


\end{document}